\documentclass[11pt]{article}
\usepackage[preprint]{acl}
\usepackage{times}
\usepackage{latexsym}
\usepackage{booktabs}
\usepackage{graphicx}
\usepackage{amsmath}
\usepackage{amssymb}
\usepackage{xcolor}
\usepackage{multirow}
\usepackage{url}

\graphicspath{{./}}

\title{Select, Compress, Reinvest:\\A Controlled Study of Visual-Token Allocation in Long-Video MLLMs}

\author{Prakhar Khatri \\
  Independent Researcher \\
  \texttt{prakharkhatri123@gmail.com}}

\begin{document}
\maketitle

\begin{abstract}
Long-video language models cannot look at every frame: an hour sampled once per second is 3{,}600 images, and a system keeps only a small fixed slice of that pool.
Which frames survive that slice is usually treated as a preprocessing detail; we test whether it should be.
Published selectors make the comparison hard because they change the frame scorer, the prompt boundary, the resolution policy, and the answering model all at once.
We hold each fixed and vary one decision at a time---selection, spatial compression, and reinvestment of the savings---across six training-free selection rules, three long-video benchmarks, and two answering models.
Selection is the largest single lever: on LongVideoBench's hour-long bin, eight query-selected frames beat sixteen uniformly spaced ones by 6.9 points, and Orthogonal Matching Pursuit---an unmodified, decades-old sparse-approximation algorithm---matches or comes within a point of every purpose-built selector we compare it against, across all three benchmarks.
Compression is close to free: halving each frame's spatial budget at fixed timestamps costs at most 0.44 points.
Reinvestment is where that budget turns back into accuracy: spending the freed tokens on twice as many compressed frames, at a measured cost no higher than the original eight, returns a further two to three points; compression only pays off once its savings are spent this way.
Along the way, an implementation bug in our own AKS baseline and a 0.07--3.74-point gap between two harnesses running the same published rules at the same budget show why these comparisons need to happen inside one controlled harness rather than across papers.
\end{abstract}

\section{Introduction}

A video language model cannot look at a whole video.
Decoded at one frame per second, a ten-minute clip is 600 images and an hour-long one is 3{,}600; a long-video system typically keeps only a small fixed slice of that pool: eight frames in some systems, thirty-two or sixty-four in others, never the whole thing.
Everything the model will ever know about the video passes through that handful of frames, so the rule that picks them is not a preprocessing detail.
It is the first and tightest bottleneck in the pipeline.

The default rule is still uniform sampling: take a fixed number of frames spread evenly across the timeline, regardless of what was asked.
We test at eight, the one budget at which a published Qwen3-VL-8B/LongCLIP baseline exists to match against (Table~\ref{tab:published}, \S\ref{sec:discussion}).
Our clearest single result is that this default is expensive.
On hour-long LongVideoBench videos, eight frames chosen for their relevance to the question beat sixteen uniformly spaced frames by 6.9 points ($p{=}.0011$), and the same pattern holds one budget lower on ten-minute videos.
Half the frames, better answers.
Spending an \emph{answerer visual-token} budget on \emph{which} frames to keep buys more than spending it on \emph{how many}; the selection stage has its own cost, which we account for in \S\ref{sec:selection} and do not include here.

Many recent systems already replace uniform sampling with query-aware retrieval, temporal coverage, diversity, or learned evidence scores~\citep{tang2025aks,sun2025mdp3,zhu2025focus,wang2026evidential,peng2026qca}, and others jointly adapt frame choice and spatial resolution~\citep{zhang2025qframe,chen2026lddr,wang2026dafs} or prune tokens once frames are inside the vision stack~\citep{vistallm2026,moprune2026}.
They report large end-to-end gains.
What they do not settle is a narrower question: holding the scorer, the frame budget, and the answering model constant, which selection rule actually supplies the most useful evidence?
Published tables cannot answer this, because those components move together.
A selector can look strong because it ships with a better encoder.
A dynamic-resolution system can look strong because it fits in more frames, not because it allocates resolution well.
Comparing two numbers from two papers compares two experiments, not two ideas.

This paper isolates the pieces.
We fix one frame scorer, one prompt boundary, one answering model, and one evaluation harness, then intervene on a single decision at a time.
\textbf{Select}: change which timestamps are chosen, holding the frame count and resolution fixed.
\textbf{Compress}: hold those timestamps fixed and shrink the spatial budget spent on each frame.
\textbf{Reinvest}: spend the recovered budget on more timestamps rather than sharper ones.
Because each step is a paired comparison on the same questions, the difference it produces is attributable to that step alone.
A fourth comparison swaps the scorer itself, to check that the resulting ordering is not an artifact of the encoder we happened to freeze.

The selection rule at the centre of these interventions is deliberately old.
Orthogonal Matching Pursuit~\citep{pati1993omp} is a greedy sparse-approximation algorithm from 1993: it repeatedly takes the candidate most correlated with what the query still needs, then projects that direction away before choosing again.
We adopt it unmodified and untuned.
That is a methodological choice rather than a concession.
An off-the-shelf rule has no hyperparameters for us to fit to these benchmarks, so a gap between it and a competing selector reflects the controlled comparison rather than unequal effort spent on our side of it.
It also sets an informative bar: an off-the-shelf rule with no video-specific machinery is about as unsophisticated as a selector can be, and how well it does is itself a measurement of how much the recent purpose-built selectors gain from their subset rules as opposed to their scorers and pipelines.

The three interventions rank consistently across the settings we tested, though they were estimated in separate contrasts rather than one factorial design, so the ordering is a synthesis rather than an identified effect ranking.
Selection is the large lever: OMP improves on uniform sampling by 5.7 to 11.8 points depending on the benchmark, and can beat uniform inputs carrying twice as many frames.
Compression is nearly free but not by itself useful: halving the per-frame budget at fixed timestamps changes accuracy by at most 0.44 points, with a post-hoc $\pm3$-point equivalence interval on pooled long videos.
Reinvestment converts that free budget into a real gain of two to three points.
None of this is universal.
The benefit depends on which model reads the frames and which benchmark asks the question, and OMP's later picks drift toward content that is visually novel but irrelevant.
We report those boundaries rather than smoothing them.

\paragraph{Contributions.}
\begin{itemize}
  \item \textbf{A matched comparison of six training-free selectors} under one scorer, one prompt boundary, one frame budget, and one answerer harness. OMP is a strong query-focused rule and stays within one point of LDDR's stage-1 selector.
  \item \textbf{Selection substitutes for frame count.} OMP significantly outperforms uniform sampling while using half as many frames, in two disjoint LongVideoBench duration bins (600 and 3600\,s) that share a benchmark, scorer, answerer, and implementation.
  \item \textbf{A decomposition of the fixed-budget allocation decision.} Roughly halving the per-frame spatial budget preserves accuracy, bounded by a two-one-sided-test interval rather than a bare null result, at a margin chosen post hoc; reinvesting the savings in additional keyframes then improves long-video QA.
  \item \textbf{The ranking survives a scorer swap.} Replacing LongCLIP with SigLIP changes 67--84\% of the selected frames yet leaves the selector ordering intact, within a test that bounds only scorer effects above roughly five points.
  \item \textbf{Aggregate accuracy can conceal implementation error.} A padding branch in our own AKS port made that baseline select global top-$k$ at every $k<32$. Correcting it moved roughly 99.5\% of selected frames yet changed LVBench accuracy by 0.07 points. We report the cross-arm overlap check that caught it.
  \item \textbf{Explicit boundaries on these claims}, from paired tests, two answerer families, three benchmarks, a residual-geometry diagnostic, and a purposive failure audit. The boundaries are uneven: the scorer swap covers one bin, the equivalence result one pooled setting, and the audit is exploratory.
\end{itemize}

\section{Related Work}

\paragraph{Query-aware frame selection.}
Training-free selectors combine three objectives in different proportions: relevance to the question, coverage of the timeline, and diversity within the chosen set.
The latter two are inherited from classical subset selection---maximal marginal relevance~\citep{carbonell1998mmr} and determinantal point processes~\citep{kulesza2012dpp}---which we also run directly as selector arms (Appendix~\ref{app:negatives}).
AKS recursively partitions the timeline to preserve coverage~\citep{tang2025aks}; FOCUS treats temporal clips as arms in a budgeted bandit~\citep{zhu2025focus}; MDP3 models relevance, list-wise diversity, and sequentiality with a DPP-based dynamic program~\citep{sun2025mdp3}; and AdaRD-Key and Adaptive Greedy pursue the same relevance--diversity tradeoff~\citep{adardkey2025,adaptivegreedy2026}.
Segment-based methods make the structure explicit: QCA assigns per-segment quotas from relevance and content deviation~\citep{peng2026qca}, while EFS forms event-like segments and refines query-relevant anchors with adaptive maximal marginal relevance~\citep{chen2026efs}.
ReQuest adds uncertainty-triggered computation and adaptive temporal suppression~\citep{kim2026request}, and query-conditioned evidential sampling learns a frame-level estimate of conditional evidence~\citep{wang2026evidential}.
These methods differ not only in the subset rule but also in scorer, candidate pool, training requirements, and frame budget, so the reported gaps between them confound the rule with its surrounding pipeline.

\paragraph{Selection with spatial adaptation.}
A second line couples frame choice to resolution.
Q-Frame pairs CLIP-based query-aware selection with multi-resolution scaling so more frames fit inside a compute limit~\citep{zhang2025qframe}; LDDR combines a linear-DPP selector with Group-DPP importance for frame retention and dynamic resolution~\citep{chen2026lddr}; DAFS extracts relevance from MLLM attention and jointly optimizes candidate-pool size and per-frame token budget~\citep{wang2026dafs}.
These systems demonstrate that joint allocation works, but a joint gain has at least three possible sources: better timestamps, better resolution policy, or simply more temporal coverage.
None of these papers separates the three, so the mechanism behind their improvements remains open.

\paragraph{Visual-token reduction.}
Token-level methods cut cost after or during visual encoding: AdaCodec learns retain, compress, and drop decisions~\citep{adacodec2026}; AdaptToken assigns group budgets from answerer uncertainty~\citep{adapttoken2026}; Vista-LLM prunes with query guidance before inference~\citep{vistallm2026}; MoPrune uses scene structure and motion to retain informative tokens~\citep{moprune2026}.
Concurrent work poses our question directly: AdaAlloc~\citep{an2026adalloc} asks how much of a fixed budget to spend on global context versus high-resolution local evidence, and answers it with a temporal-grounding and memory pipeline that assembles a hybrid low-resolution/high-resolution input. No preprint or code was publicly available at the time of writing, so we describe it from its project-page abstract and draw no comparison against our numbers.
We ask a coarser but experimentally separable question that sits upstream of all of them: before any token-level pruning, does a long-video answerer gain more from sharper retained frames or from additional moments?
Our target is the evidence such an allocator needs rather than the allocator itself.

\section{Method: Three Controlled Interventions}
\label{sec:method}

\begin{figure*}[t]
\centering
\includegraphics[width=\textwidth]{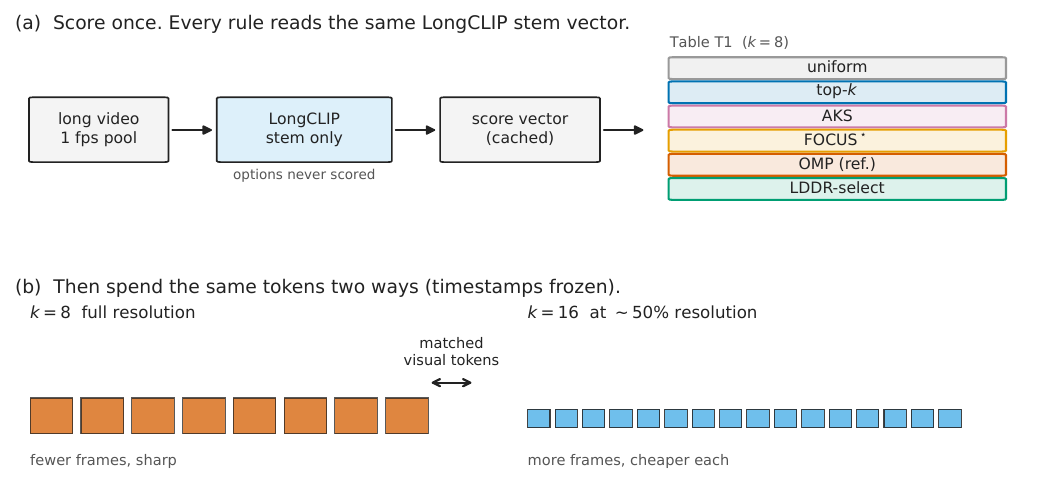}
\caption{Controlled protocol.
(a) A 1\,fps candidate pool is encoded once with LongCLIP using only the question stem; every $k{=}8$ selector consumes the same cached embeddings and scores.
(b) Selected timestamps are then frozen while spatial budget is reduced.
(c) The saved visual tokens are reinvested by comparing eight full-resolution frames with sixteen compressed frames at equal or lower measured token cost.
FOCUS$^\star$ denotes a replay of the published selection schedule on dense LongCLIP scores, not its original budgeted ITM scorer.}
\label{fig:protocol}
\end{figure*}

\subsection{The experimental unit}

Every comparison in this paper is paired.
The unit is a single benchmark question answered twice, once under each of two input policies, with everything else held constant.
This lets us count the questions each policy wins rather than compare two accuracy numbers that may differ for unrelated reasons.
The three interventions differ only in what the two policies are allowed to change:
\textbf{selection} changes the timestamps while holding the frame count, scorer, and resolution fixed;
\textbf{compression} holds the timestamps fixed and changes the per-frame spatial budget;
\textbf{reinvestment} spends the budget saved by compression on additional timestamps.
Measuring the marginal value of one decision at a time is what distinguishes this from a comparison of complete systems whose components all differ at once.
A fourth intervention sits outside that sequence and tests the design itself: replacing the scorer while holding the selection rule, budget, and answerer fixed asks whether the ordering the first three produce is a property of the rules or of one encoder (\S\ref{sec:scorer}).

\subsection{One scorer, shared by every rule}

Videos are decoded at 1\,fps.
Each candidate frame and the question stem are encoded once with LongCLIP~\citep{zhang2024longclip}, and the resulting embeddings and stem similarities are cached.
Every selector then reads the same cache, so no rule benefits from a better encoder than another.

Answer options never enter selection.
That is a convention rather than a neutral choice, so we measure what it costs.
Scoring against the fused question-and-options string instead of the stem alone changes 41.9\% of the frames top-$k$ selects on LongVideoBench-600\,s and 53.3\% of those OMP selects; OMP is the more sensitive of the two because each pick is chosen to explain a residual component of the query vector, so a perturbation of that vector compounds through the greedy chain.
Those are also better frames, though not because of the options themselves.
On that bin, with the fused query, OMP reaches .6699 against .6311 for the stem (${+}3.88$ points, $p{=}.033$, 33 rescued against 17 broken) and top-$k$ reaches .6408 against .6068 (${+}3.40$, $p{=}.087$).
A pre-registered replication on the 3600\,s bin gives the same picture: OMP reaches .5691 against .5461 for the stem (${+}2.30$ points, $n{=}564$, 46 rescued against 33 broken, $p{=}.18$), with the fused query moving roughly 69\% of the selected frames.
Neither long bin resolves the shift on its own at these sample sizes, but the sign and magnitude agree across two disjoint video sets, so we read the prompt boundary as a level shift affecting long videos generally rather than a property of the 600\,s bin.
A control arm isolates what the answer options contribute.
Replacing each item's options with those of a different video---unrelated in content, matched on length, and perturbing 54\% of OMP's picks against the real options' 53\%---reaches .6529, recovering 56\% of the gain; the item's own options add a further 1.70 points beyond that, which is not significant ($p{=}.42$).
Neither the control against the stem (${+}2.18$, $p{=}.23$) nor the fused query against the control separates at this sample size, so we do not claim a mechanism.
What the control does rule out is that the effect belongs to the answer set: a scorer query built from the wrong options moves accuracy about as far as one built from the right ones.
We nonetheless score with the stem alone, following the official AKS implementation, which passes the question without candidates to its CLIP and BLIP branches~\citep{tang2025aks}.
Every selector in this paper is therefore evaluated in a deliberately conservative configuration, and the gap above is the price of that choice rather than a hazard it avoids.
The consequence for reading the literature is the substantive one: the text handed to the scorer is an uncontrolled axis worth roughly two to four accuracy points across the two long bins---and, on the evidence above, one that does not require the substituted text to be relevant---so a comparison inconsistent about it is measuring the prompt rather than the selector.

\subsection{Selection rules}

The matched comparison contains uniform sampling, cosine top-$k$, AKS~\citep{tang2025aks}, FOCUS$^\star$~\citep{zhu2025focus}, OMP, and LDDR-select~\citep{chen2026lddr}.
Every rule here consumes our LongCLIP stem scores, but that is native only for LDDR-select.
AKS and FOCUS both score with BLIP-ITM in their published form, so both are reproductions of a subset rule under a substituted scorer, and LDDR-select is the only row evaluated with the encoder its authors used, an advantage to keep in mind when reading its position in Table~\ref{tab:t1}.
FOCUS$^\star$ carries a star for a second reason: it replays the published clip-bandit schedule against the dense LongCLIP vector rather than the original budgeted online ITM process, which is the substance of the method; Appendix~\ref{app:focus} states exactly what this does and does not test.
LDDR-select is the stage-1 Linear-DPP selector (LD) only, not the full dynamic-resolution pipeline; it is a published, separately tabulated configuration of that system rather than a truncation we invented.

Let $E\in\mathbb{R}^{N\times d}$ contain L2-normalized frame embeddings and let $q_0$ be the normalized stem embedding.
At each step OMP takes the frame most correlated with the current residual, then subtracts the span of everything already selected from the query:
\begin{equation}
\begin{aligned}
b_r &= \operatorname*{arg\,max}_{i\notin B_{r-1}} e_i^\top q_{r-1},\\
q_r &= q_0-\operatorname{Proj}_{\operatorname{span}\{e_j:j\in B_r\}}(q_0),
\end{aligned}
\label{eq:omp}
\end{equation}
where $B_r=B_{r-1}\cup\{b_r\}$.
In its original signal-processing setting this is sparse reconstruction: the residual shrinks as the selected atoms explain more of the target.
Whether that interpretation survives in a contrastive image--text embedding space is an empirical question, not an assumption, and \S\ref{sec:mechanism} tests it directly.
What the projection reliably does here is suppress directions already covered by earlier picks.

\subsection{Spatial budget and reinvestment}

For the fixed-timestamp intervention, frames are resized before reaching the model's processor, whose own internal resize is disabled for Qwen3-VL so that our resize is the only one applied.
The main compressed arm uses a residual-proportional schedule targeting a mean spatial fraction of about 0.53, which we label D@53.

Two controls test what the schedule's \emph{shape} contributes.
A flat split at the same mean fraction asks whether OMP's ranking is a useful priority for spending pixels.
A reconstruction of LDDR's stage-2 Group-DPP importance, applied to identical OMP-8 timestamps, asks whether a different importance signal can do better than either.

The reinvestment arm selects $k{=}16$ timestamps and applies an approximately 50\% per-frame cap, a different quantity from the 0.53 mean used when timestamps stay fixed.
Because the comparison only means something if the two arms really cost the same, we audit tokens empirically rather than assuming the resize ratio transfers: a script reproduces Qwen's smart-resize rules from each video's own dimensions.
The sixteen-frame arm consumes 0.996 and 0.984 times the tokens of the eight-frame full-resolution arm in the 600\,s and 3600\,s LongVideoBench bins.
Both ratios are below one, so the LongVideoBench reinvestment result is conservative: the winning arm is also the cheaper one.
Video-MME and LVBench use the same resize rule, but we did not run a separate token audit on them and do not claim audited parity there.

\section{Evaluation Setup}
\label{sec:setup}

We evaluate on LongVideoBench~\citep{wu2024longvideobench} (the complete official validation set, $n{=}1337$), Video-MME~\citep{fu2025videomme} ($n{=}2700$), and LVBench~\citep{lvbench2025} ($n{=}1549$).
Tables abbreviate the first two as LVB and V-MME where column width requires it.
LongVideoBench is additionally stratified into 15, 60, 600, and 3600\,s duration bins, which is what makes the duration analysis in \S\ref{sec:selection} possible.
The primary answerer is Qwen3-VL-8B-Instruct~\citep{qwen3vl2025}, run through lmms-eval~\citep{zhang2024lmmseval} with greedy decoding, temperature zero, and subtitles disabled.
InternVL3-2B and InternVL3-8B~\citep{zhu2025internvl3} provide a second open-model family, and GPT-5-mini checks selected contrasts on the same saved timestamps.

\begin{table}[t]
\centering
\small
\begin{tabular}{@{}lp{0.58\columnwidth}@{}}
\toprule
Benchmarks & LongVideoBench (1337), Video-MME (2700), LVBench (1549) \\
Primary answerer & Qwen3-VL-8B, greedy decoding, subtitles off \\
Transfer checks & InternVL3 2B/8B; GPT-5-mini \\
Shared scorer & LongCLIP, question stem only, 1\,fps pool \\
Scorer control & SigLIP-so400m, identical protocol (\S\ref{sec:scorer}) \\
Frame budgets & $k{=}8$ for selection; $k{=}16$ for reinvestment \\
\bottomrule
\end{tabular}
\caption{Evaluation grid. Main claims concern paired accuracy differences under matched inputs, not absolute leaderboard position.}
\label{tab:setup}
\end{table}

\paragraph{Statistics.}
Accuracy comparisons use exact two-sided McNemar tests~\citep{mcnemar1947} on paired correctness outcomes, which is the appropriate test when the same question is answered under both policies.
Compression claims need more than a non-significant McNemar result, because failing to detect a difference is not evidence that none exists.
Where we claim two arms are interchangeable we therefore report two one-sided tests (TOST) with 90\% confidence intervals~\citep{schuirmann1987}, which bound the effect rather than merely failing to find it.
The $\pm2$, $\pm3$, and $\pm4$ percentage-point margins were chosen after seeing the runs rather than preregistered; we therefore report both the narrowest margin the data support and the adjacent margin they do not.
Tests are contrast-specific and uncorrected for multiplicity.

\section{Results}
\label{sec:results}

\subsection{Choosing frames beats having more of them}
\label{sec:selection}

The most direct evidence that selection matters is that it substitutes for frame count.
On 3600\,s LongVideoBench videos, OMP with eight frames reaches .5461 while uniform sampling with sixteen frames reaches .4770, a 6.9-point gain on half the input ($p{=}.0011$).
The same pattern holds one budget higher: on 600\,s videos, OMP with sixteen frames reaches .6578 against .6044 for uniform sampling with thirty-two, a 5.3-point gain at half the frame count ($p{=}.018$).
Both contrasts compare a policy that is cheaper \emph{in answerer visual tokens} against a more expensive one, and the cheaper policy wins, which is a stronger statement than any equal-budget comparison can make.
This accounting covers the answerer only: producing the selected set additionally requires decoding the video at 1\,fps and encoding the whole pool with LongCLIP, a cost we do not measure and which a single-query-per-video deployment would not amortise.
Doubling the frame count is the obvious way to spend more compute on a long video, and in these two LongVideoBench bins it is the worse way.

\begin{table*}[t]
\centering
\small
\begin{tabular}{lccc}
\toprule
Method & LongVideoBench & Video-MME & LVBench \\
\midrule
Uniform & .5654 ($-$5.69) & .5637 ($-$5.85) & .3454 ($-$11.81) \\
Top-$k$ & .6028 ($-$1.95) & .5704 ($-$5.18) & .4319 ($-$3.16) \\
\textbf{OMP} & .6223 & \textbf{.6222} & .4635 \\
AKS & .5916 ($-$3.07) & .6059 ($-$1.63) & .4287 ($-$3.48) \\
FOCUS$^\star$ & .5819 ($-$4.04) & .5578 ($-$6.44) & .3983 ($-$6.52) \\
LDDR-select & \textbf{.6320} (${+}0.97$) & .6193 ($-$0.29) & \textbf{.4693} (${+}0.58$) \\
\bottomrule
\end{tabular}
\caption{Matched selector comparison at $k{=}8$ with Qwen3-VL-8B. Cells are accuracy; parentheses give the difference from OMP in percentage points. FOCUS$^\star$ is the controlled schedule replay described in Appendix~\ref{app:focus}; LDDR-select is stage~1 only.}
\label{tab:t1}
\end{table*}

Table~\ref{tab:t1} places that result in the wider comparison, with all six rules reading the same cached scores.
OMP improves on uniform sampling by 5.69 points on LongVideoBench, 5.85 on Video-MME, and 11.81 on LVBench, and the paired evidence is decisive on Video-MME ($p{=}5.1\times10^{-10}$), LVBench ($p{=}9.5\times10^{-17}$), and the LongVideoBench 3600\,s bin ($p{=}5.9\times10^{-4}$).
LDDR-select is the one rule that keeps pace, staying within a point of OMP on all three benchmarks with no detectable paired difference on Video-MME or LVBench ($p{=}.70$ and $.57$).

\begin{figure}[t]
\centering
\includegraphics[width=\columnwidth]{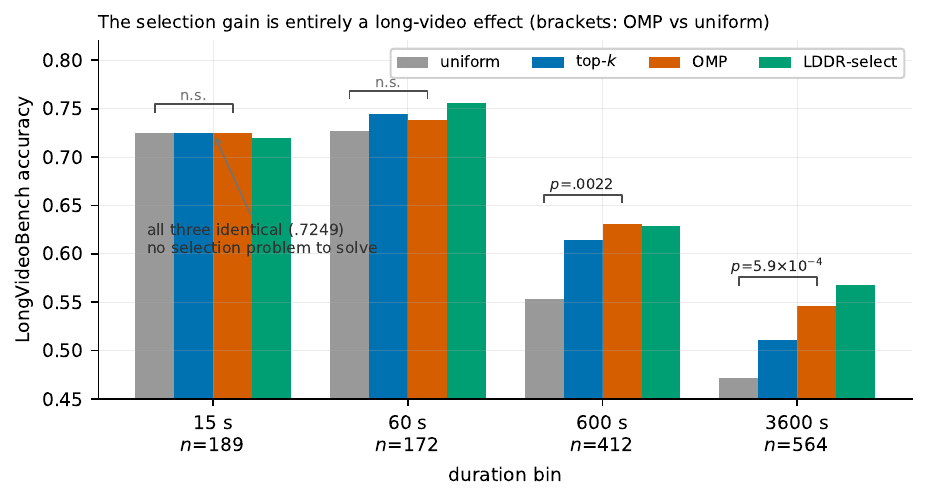}
\caption{OMP's gain over uniform sampling appears only once the 1\,fps candidate pool outgrows the eight-frame budget. At 15\,s, uniform, top-$k$, and OMP have identical accuracy (.7249); the gain is 7.8 points at 600\,s and 7.5 at 3600\,s. The effect switches on with pool size rather than growing with duration.}
\label{fig:longvideo}
\end{figure}

The duration split explains when any of this matters.
On 15\,s clips, uniform sampling, top-$k$, and OMP return exactly the same accuracy (.7249): eight frames out of fifteen candidates cover the video regardless of how they are picked, so there is nothing for a selector to do.
The gains appear once the candidate pool outgrows the budget: 7.8 points at 600\,s ($p{=}.0022$) and 7.5 points at 3600\,s ($p{=}5.9\times10^{-4}$).
Note that the gain does not keep growing after that: 7.5 at 3600\,s is if anything slightly below the 7.8 at 600\,s, so this is a threshold that switches on when the pool exceeds the budget, not a gradient in duration.
The duration bins also differ in content and question composition, so this is an observational contrast between bins; isolating pool size from duration would require a controlled truncation experiment that we did not run.

Which \emph{kind} of rule is needed is more benchmark-dependent than the headline suggests.
On Video-MME, top-$k$ and FOCUS$^\star$ do not separate detectably from uniform sampling, while OMP, LDDR-select, and AKS do---AKS by 4.22 points ($p{=}4.0\times10^{-6}$); on LongVideoBench, OMP's advantage over plain top-$k$ is small and often undetectable within individual bins.
The supported reading is narrow and worth stating precisely: combining query relevance with suppression of already-covered directions is consistently strong in this harness, whereas the value of any individual simpler heuristic does not transfer across benchmarks on its own.
Whether that ordering is a property of the rules or of LongCLIP is the subject of \S\ref{sec:scorer}.

\subsection{The ranking survives a scorer swap}
\label{sec:scorer}

Every number in Table~\ref{tab:t1} comes from a single frozen scorer, which leaves open whether the ordering reflects the selection rules or LongCLIP's particular geometry.
We test that directly, replacing LongCLIP with SigLIP-so400m~\citep{zhai2023siglip} and holding the prompt boundary, frame budget, answerer, and harness fixed.
Both scorers' arms were re-run together in one environment rather than comparing fresh SigLIP numbers against the banked LongCLIP results, so every contrast below is paired on identical items.

The swap is a substantial intervention rather than a cosmetic one.
On the 600\,s bin the two scorers agree on only 1.32 of eight frames for OMP and 2.62 for top-$k$, which is seven to fourteen times the chance overlap for these pool sizes: both are selecting rather than scattering, yet 67 to 84\% of what the answerer sees changes.

\begin{table}[t]
\centering
\small
\begin{tabular}{lccc}
\toprule
Selector & LongCLIP & SigLIP & SigLIP $-$ LongCLIP \\
\midrule
Uniform & \multicolumn{2}{c}{.5534} & --- \\
Top-$k$ & .6068 & .6068 & ${+}0.00$ ($p{=}1.00$) \\
OMP & .6311 & .6408 & ${+}0.97$ ($p{=}.73$) \\
\bottomrule
\end{tabular}
\caption{Scorer swap on LongVideoBench-600\,s ($n{=}412$), all arms run together and paired on identical items. Uniform sampling is scorer-independent and appears once. The final column is a paired test of whether each selector's gain depends on the scorer.}
\label{tab:scorer}
\end{table}

Table~\ref{tab:scorer} shows the ordering surviving intact.
Uniform sampling is beaten by top-$k$ and top-$k$ by OMP under both scorers, and OMP's advantage over uniform is significant in each case ($p{=}.0024$ under LongCLIP, $p{=}.00026$ under SigLIP).
Discordant counts for these contrasts are $b{=}63,c{=}41$ and $b{=}69,c{=}37$ under LongCLIP and $b{=}60,c{=}38$ and $b{=}65,c{=}29$ under SigLIP, for top-$k$ and OMP respectively.
Neither selector's gain depends detectably on which scorer produced it, but the test is not powerful: with roughly 134 discordant pairs the 90\% interval on the OMP scorer effect spans about $[-3.6,+5.6]$ points, so this bounds large scorer effects rather than establishing their absence.
The 60\,s bin adds nothing either way: no contrast there reaches significance (the largest is ${+}3.49$ points at $p{=}.31$), which is the expected behaviour when eight frames already cover 22\% of a median 36-frame pool.

One reading must be resisted.
OMP beats top-$k$ significantly under SigLIP (${+}3.40$, $p{=}.049$) and not under LongCLIP (${+}2.43$, $p{=}.33$), which invites the conclusion that OMP suits SigLIP better.
The interaction test rejects that: the difference of the two gains is 0.97 points ($b{=}48$, $c{=}43$, $p{=}.68$).
That estimate coincides with the OMP row of Table~\ref{tab:scorer} only because top-$k$ has identical marginal accuracy under both scorers; the two are different tests over different per-item patterns, which is why their $p$-values differ.
A significant result beside a non-significant one is not evidence that the two differ, and Appendix~\ref{app:qtype} records the same error made on question categories.

The claim this supports is bounded.
One alternative scorer on two duration bins does not establish scorer-invariance in general, and the two encoders share a contrastive image--text objective, so a scorer built on a different principle---an ITM head, or MLLM attention---could still reorder the table.
What it weakens---without ruling out---is the worry that Table~\ref{tab:t1} is an artifact of LongCLIP: the ordering holds for the three rules tested when most of the selected frames are replaced.

\subsection{Half the pixels change almost nothing}
\label{sec:compression}

Table~\ref{tab:t2} holds the selected timestamps fixed and changes only how many pixels each frame is worth.
Across the three Qwen benchmark aggregates, moving to the compressed arm shifts accuracy by at most 0.44 points in either direction.
A uniform-timestamp control on pooled LongVideoBench long videos moves by 0.31 points ($p{=}.784$), which is consistent with the stability being a property of the compression rather than of OMP-selected frames specifically; a non-significant control is not itself an equivalence result.

\begin{table}[t]
\centering
\small
\resizebox{\columnwidth}{!}{%
\begin{tabular}{lccc}
\toprule
Answerer / benchmark & Full & Compressed & $\Delta$ (pp) \\
\midrule
Qwen / LongVideoBench & .6223 & .6253 & ${+}0.30$ \\
Qwen / Video-MME & .6222 & .6178 & $-0.44$ \\
Qwen / LVBench & .4635 & .4674 & ${+}0.39$ \\
GPT-5-mini / LVBench & .4900 & .4926 & ${+}0.26$ ($p{=}.84$) \\
Uniform / LVB long pool & .5061 & .5092 & ${+}0.31$ ($p{=}.784$) \\
\bottomrule
\end{tabular}}
\caption{Fixed-timestamp spatial compression. The Qwen rows use the saved approximately half-budget arm; the GPT row uses a pixel-ratio control and is not a Qwen token count. The uniform control uses an approximately 46\% realized per-frame budget.}
\label{tab:t2}
\end{table}

Small differences are easy to over-read, so we bound one of them properly.
On the pooled LongVideoBench long bins ($n{=}976$), the compressed OMP arm is 0.72 points \emph{higher} than full resolution, with a 90\% confidence interval of $[-0.60,+2.03]$.
That interval fits inside a $\pm3$-point margin (TOST $p{=}.0022$) but narrowly fails at $\pm2$ ($p{=}.054$), so $\pm3$ is the honest statement.
InternVL3-8B supports a stronger version under a more aggressive intervention: cutting 24 tiles to 8 moves pooled accuracy by $-0.10$ points with a 90\% interval of $[-1.66,+1.45]$, which clears a $\pm2$-point margin ($p{=}.022$).
The remaining LVBench and Video-MME differences in Table~\ref{tab:t2} are descriptive stability checks; we do not present them as equivalence results.

\begin{figure}[t]
\centering
\includegraphics[width=\columnwidth]{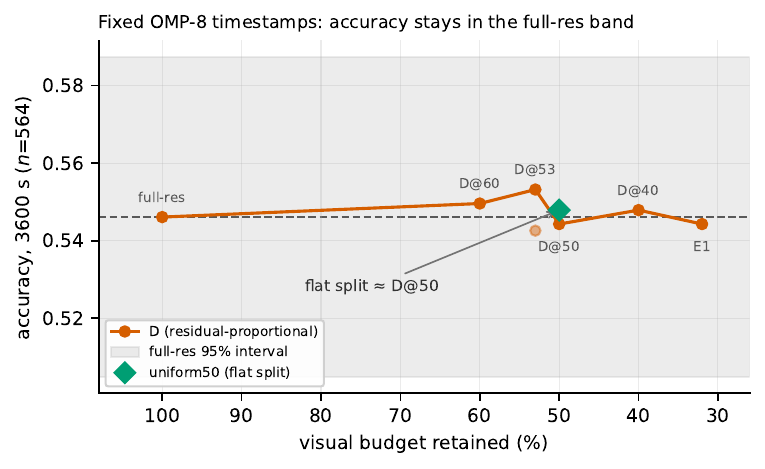}
\caption{LongVideoBench 3600\,s accuracy on fixed OMP-8 timestamps across spatial budgets. The flat 50\% control and the residual-proportional schedule are indistinguishable, so OMP rank alone does not provide a useful resolution policy.}
\label{fig:claim-a}
\end{figure}

Compression being cheap does not mean the allocation \emph{shape} is irrelevant, and our controls separate the two.
The residual-proportional schedule does not beat a flat split at the same mean budget (Figure~\ref{fig:claim-a}), so OMP's ranking---useful as it is for choosing frames---did not improve accuracy detectably when used to distribute pixels among them. We ran no equivalence test on that contrast, so this is a failure to detect a difference rather than evidence of none.
A reconstruction of LDDR's stage-2 Group-DPP importance does better, improving on the flat split by 1.84 points on the pooled long bins ($p{=}.0198$, 95\% CI $[+0.37,+3.32]$).
That contrast comes from our own same-environment reconstruction rather than the authors' released implementation, and its 3600\,s component is not individually significant, so we treat it as indicative.
The conclusion is therefore three-part and narrower than ``resolution does not matter'': coarse compression is close to free, a naive priority-proportional rule adds nothing, and a better-designed importance signal can add something.

\subsection{Spending the savings on more frames}
\label{sec:reinvest}

The direction of this intervention is not new.
\citet{chen2026lddr} report fixed-resolution ablations under a single total token budget in which 1024 tokens per frame reach 59.76 on LongVideoBench, 512 reach 61.03, and 256 fall back to 59.31: halving the per-frame budget doubles the frames that fit, and accuracy improves by 1.27 points before declining again.
What that ablation does not establish is whether the parity it assumes actually holds, whether the difference is distinguishable from noise, or whether the effect needs their selector.
We address those three questions, so the contribution here is an audited and paired replication rather than a new observation.

If half the per-frame budget is recoverable at no cost, the question becomes what to buy with it.
Table~\ref{tab:t3} spends it on temporal coverage, comparing eight full-resolution frames against sixteen compressed ones at equal or lower measured token cost.
LongVideoBench improves by 2.24 points overall and 2.36 on the pooled 600 and 3600\,s bins ($p{=}.0346$), with almost identical positive effects in each.
LVBench gains 3.04 points ($p{=}.0009$), Video-MME 1.56, and GPT-5-mini on LVBench 2.39 ($p{=}.039$).

\begin{table}[t]
\centering
\small
\resizebox{\columnwidth}{!}{%
\begin{tabular}{lccc}
\toprule
Answerer / benchmark & $k{=}8$ full & $k{=}16$ compressed & $\Delta$ (pp) \\
\midrule
Qwen / LongVideoBench & .6223 & .6447 & ${+}2.24$ \\
Qwen / LVB long pool & .5820 & .6055 & ${+}2.36$ ($p{=}.0346$) \\
Qwen / Video-MME & .6222 & .6378 & ${+}1.56$ \\
Qwen / LVBench & .4635 & .4939 & ${+}3.04$ ($p{=}.0009$) \\
GPT-5-mini / LVBench & .4900 & .5139 & ${+}2.39$ ($p{=}.039$) \\
\bottomrule
\end{tabular}}
\caption{Reinvestment at equal or lower visual-token cost on the audited LongVideoBench bins. The compressed arm approximately doubles temporal coverage; Video-MME and LVBench use the same resize rule without a separate token audit.}
\label{tab:t3}
\end{table}

\begin{figure}[t]
\centering
\includegraphics[width=0.80\columnwidth]{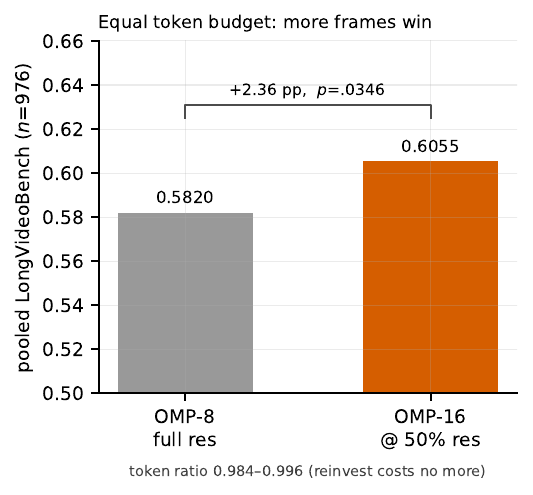}
\caption{LongVideoBench 600\,s and 3600\,s pooled: sixteen compressed OMP frames outperform eight full-resolution OMP frames at a measured token ratio no greater than one.}
\label{fig:claim-b}
\end{figure}

We cannot show that this requires OMP.
To test it we ran all four cells of the selector-by-budget design---uniform and OMP, each at eight full-resolution and sixteen compressed frames---within a single environment, so the interaction is measured without any cross-environment comparison.
Under uniform sampling the eight-to-sixteen intervention gives ${+}1.13$ points ($p{=}.3673$); under OMP it gives ${+}2.56$.
The interaction between them is 1.43 points and does not reach significance ($p{=}.2614$), with OMP gaining on 110 items where uniform does not and uniform gaining on 93 where OMP does not.
It does split by duration: essentially zero at 600\,s (${+}0.00$, $p{=}1.000$) and ${+}2.48$ at 3600\,s ($p{=}.1706$), which is the direction expected if selection matters more as the candidate pool grows, but neither bin is individually significant.
This is evidence against a \emph{large} OMP-specific interaction rather than proof the two are identical; with 203 discordant pairs the test excludes only large effects.
The defensible claim is that temporal reinvestment works with OMP, points the same way under uniform sampling, and that its selector-specificity remains open at roughly the one-point scale.

\subsection{What the later picks actually add}
\label{sec:mechanism}

OMP's residual trace turns out not to behave the way its signal-processing origin predicts. We offer the following as a hypothesis consistent with that discrepancy, not as a demonstrated mechanism: nothing here intervenes on residual geometry, so the account below is observational.
On both LongVideoBench long bins, the residual norm is 0.972 of its starting value after a single pick and 0.967 after eight, and does not fall further by sixteen.
Almost none of the query is ever reconstructed.
What does change is the residual's correlation with the frame being selected, which drops from .233 at pick one to .004 at pick eight and to roughly zero by pick sixteen.
The query-directed signal is exhausted early.

\begin{figure*}[t]
\centering
\includegraphics[width=\textwidth]{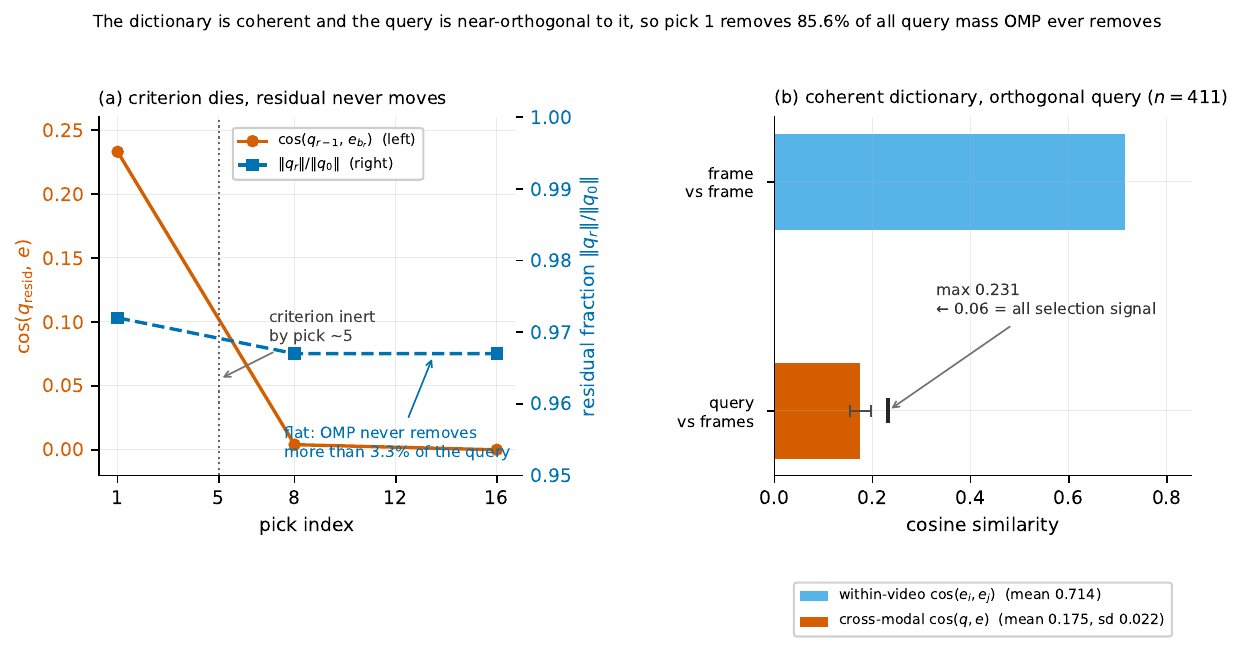}
\caption{LongCLIP geometry on LongVideoBench-600\,s ($n{=}411$).
(a) Residual correlation vanishes while residual norm changes little; the line connects logged picks 1, 8, and 16 and is not a dense trace.
(b) Within-video frames occupy a coherent cone, while the text query is nearly orthogonal to the frame dictionary. In this space, OMP mainly suppresses directions already represented by earlier selections.}
\label{fig:residual}
\end{figure*}

Yet those later frames are not junk: they keep a cosine near .20 with the original query.
The picture is a text query that sits almost orthogonal to the \emph{dominant directions} of the frame-embedding cone---the modality gap of contrastive image--text encoders~\citep{liang2022modality}, visible directly in Figure~\ref{fig:residual}b---so with an effectively low-rank frame dictionary, eight greedy picks reach only those dominant directions and OMP cannot reduce the residual much whatever it selects.
What it can do is avoid repeating itself.
Late picks add little new query direction while adding new chances for the answerer to encounter a relevant moment. That is consistent with the reinvestment result, but the two are co-occurring measurements; we did not test mediation.

Coverage helps only while it stays relevant, and a failure audit shows where that breaks.
An exploratory visual inspection of 93 OMP failures at $k{=}8$ across the two long bins found one dominant pattern: off-topic novelty, in 20 of 41 inspected 600\,s cases and 35 of 52 inspected 3600\,s cases, where several of the eight picks went to visually distinctive but question-irrelevant material: unrelated chapters, title cards, dark transitions.
The mirror-image failure also appeared: on sequence-of-scenes and cross-scene tracking questions, dense top-$k$ sampling clustered on one relevant moment and missed another required scene, which is exactly where OMP's spread paid off.
Direct $k{=}8$ versus $k{=}16$ results by category are mixed, so this supports a mechanism hypothesis rather than a rule about which question types need diversity.
It does point clearly at a design target: a relevance floor, so that spread is explored within plausible evidence regions instead of across the whole timeline.

\subsection{The same frames do not help every model equally}
\label{sec:transfer}

Table~\ref{tab:internvl} hands the identical OMP and uniform timestamps to InternVL3, changing the answerer and nothing else.
The selection gain replicates on LongVideoBench and LVBench at both model sizes.
Video-MME does not cooperate: the 2B model benefits at every duration, while the 8B model benefits only on short videos and is flat on medium and long ones despite receiving demonstrably different frames.
Capacity alone does not account for it, since the same 8B model gains 10.14 points on LVBench; a capacity-by-benchmark interaction remains possible and one model family cannot separate the two.

\begin{table}[t]
\centering
\small
\resizebox{\columnwidth}{!}{%
\begin{tabular}{lrrr}
\toprule
Benchmark & $n$ & IVL3-2B $\Delta$ & IVL3-8B $\Delta$ \\
\midrule
Video-MME short & 900 & ${+}6.00$ ($p{=}1.2\times10^{-4}$) & ${+}5.44$ ($p{=}1.9\times10^{-4}$) \\
Video-MME medium & 900 & ${+}3.33$ ($p{=}.046$) & ${+}0.33$ ($p{=}.891$) \\
Video-MME long & 900 & ${+}4.22$ ($p{=}.0062$) & ${+}0.11$ ($p{=}1.000$) \\
LVBench & 1549 & ${+}9.04$ ($p{=}7.2\times10^{-11}$) & ${+}10.14$ ($p{=}1.2\times10^{-13}$) \\
LongVideoBench & 1337 & ${+}4.71$ ($p{=}1.6\times10^{-4}$) & ${+}6.51$ ($p{<}10^{-4}$) \\
\bottomrule
\end{tabular}}
\caption{OMP minus uniform accuracy on identical timestamps and tile budgets with InternVL3. Deltas are percentage points.}
\label{tab:internvl}
\end{table}

The allocation results transfer more smoothly.
On pooled LongVideoBench long videos, InternVL3-8B holds its accuracy through a threefold tile reduction ($-0.10$ points, $p{=}1.000$) and gains 3.38 points when the same total tile budget is spread from 8 uniformly sampled frames to 24 ($p{=}.0099$), an effect concentrated in the 600\,s bin.
Both halves of the select--compress--reinvest story therefore survive a change of answerer, but their magnitudes do not: selection and allocation effects are properties of an answerer--benchmark pair, not constants of a timestamp set.

\section{Discussion}
\label{sec:discussion}

\paragraph{The three decisions are not equally worth optimizing.}
Under a fixed visual-token budget, our interventions rank consistently.
Choosing frames is the large lever: a small query-focused set beats uniform sampling, sometimes even when the uniform input carries twice as many frames.
Compressing those frames is close to free but buys nothing on its own; it changes few predictions in every Qwen and InternVL setting we tested.
Its accuracy value lies in what the savings are spent on, and spending them on temporal coverage returns a further two to three points; compression may still be worth it for latency or memory even when the tokens are not reinvested.
The practical ordering is therefore to fix selection first, treat resolution as the slack variable, and reinvest rather than pocket the difference.

\paragraph{An off-the-shelf pursuit algorithm is competitive with purpose-built selectors.}
The most uncomfortable observation in Table~\ref{tab:t1} is which rules are close together.
OMP has no video-specific machinery and no tuned hyperparameters, and nevertheless leads AKS, FOCUS$^\star$, and top-$k$ on all three benchmarks while sitting within a point of LDDR's stage-1 selector.
We do not read this as evidence that recent selectors are poorly designed.
We read it as evidence about where their gains live.
Once the scorer, prompt boundary, frame budget, and answerer are held fixed, the room between subset-selection rules is smaller than the distance from uniform sampling to plain top-$k$ on LVBench alone (4.06 points against 8.65).
On LongVideoBench the two are comparable (4.04 against 3.74), and on Video-MME the between-rule spread dwarfs it (5.18 against 0.67).
Video-MME is the sharpest case: there top-$k$ gains only 0.67 points over uniform while OMP, LDDR-select, and AKS each gain more than four, so the choice of \emph{which} rule matters far more than the choice to be query-aware at all.
The defensible generalization is therefore not ``subset rules barely matter'': on one of three benchmarks the between-rule span is the smaller quantity, and on the other two it is not. We do not apportion variance between scorer, pipeline, implementation, and answer parsing, so this compares two spans rather than attributing effects to components.
The corollary for evaluation practice is concrete: a selector comparison is only informative if the scoring stage is shared, and a paper reporting a new rule should include a classical pursuit or greedy baseline on its own scores before claiming the rule is what helped.

\begin{table}[t]
\centering
\small
\resizebox{\columnwidth}{!}{%
\begin{tabular}{lcccccc}
\toprule
& \multicolumn{2}{c}{Video-MME} & \multicolumn{2}{c}{LongVideoBench} & \multicolumn{2}{c}{LVBench} \\
\cmidrule(lr){2-3}\cmidrule(lr){4-5}\cmidrule(lr){6-7}
Rule & pub. & ours & pub. & ours & pub. & ours \\
\midrule
\emph{uniform (absolute)} & 57.19 & 56.37 & 54.53 & 56.54 & 28.08 & 34.54 \\
\midrule
AKS$^\dagger$ & ${+}4.29$ & ${+}4.22$ & ${+}5.01$ & ${+}2.62$ & ${+}12.07$ & ${+}8.33$ \\
FOCUS$^{\star\dagger}$ & ${+}1.70$ & $-0.59$ & ${+}5.00$ & ${+}1.65$ & ${+}3.49$ & ${+}5.29$ \\
LDDR-select & ${+}5.96$ & ${+}5.56$ & ${+}8.90$ & ${+}6.66$ & ${+}14.98$ & ${+}12.39$ \\
OMP & --- & ${+}5.85$ & --- & ${+}5.69$ & --- & ${+}11.81$ \\
\bottomrule
\end{tabular}}
\caption{Two matched-condition harnesses running the same rules. Published columns are \citet{chen2026lddr}, Table~1, Qwen3-VL-8B at eight frames with a LongCLIP encoder---the same answerer family, budget, and encoder as ours---where LDDR-select is their stage-1 row. The first line gives absolute uniform-sampling accuracy; all other cells are gains over the uniform row of the same column. $^\dagger$ marks rules we run on a substituted scorer (both score with BLIP-ITM natively), and $^\star$ marks the FOCUS schedule replay of Appendix~\ref{app:focus}. OMP has no published counterpart, and MDP3 and Q-Frame appear in their table but not ours (\S\ref{sec:limitations}).}
\label{tab:published}
\end{table}

\paragraph{Two matched harnesses disagree by as much as the effects they measure.}
LDDR is the one prior study that standardises every baseline to a single encoder, and it reports a cell matched to ours on answerer family, frame budget, and encoder.
Table~\ref{tab:published} places that cell beside our own.
Our gains are smaller in eight of the nine comparable cells, by 0.07 to 3.74 points.
We do not read this as either harness being wrong.
The uniform baselines themselves differ---most starkly on LVBench, 28.08 against our 34.54---so the two harnesses are not measuring from the same floor. A lower floor plausibly leaves more room for a selector to gain, though we did not test that and offer it as a hypothesis rather than an explanation.
Prompt template, candidate-pool construction, decoding path, and answer parsing all differ and none of them is specified tightly enough in either paper for a reader to reconcile the two.
We also find no clean pattern in the residuals, and in particular no advantage to keeping a rule's native encoder: on Video-MME the closest reproduction is AKS$^\dagger$, within 0.07 points of the published gain despite a substituted scorer, while LDDR-select---the one rule we run on its authors' own encoder---reproduces closest on LongVideoBench (2.24) and LVBench (2.59).

\paragraph{A baseline port that silently reproduced top-$k$.}
Our first implementation of AKS ended in a padding branch that is absent from the authors' released code.
Its recursive partition gives each leaf at depth $d$ a quota of $\lfloor k/2^{d}\rfloor$ frames, so at $k{=}8$ with the published depth of five every leaf is allotted zero, the recursion returns nothing, and our pad filled the entire selection from global cosine top-$k$.
For every benchmark in an earlier draft of this paper, the AKS row was a duplicate of the top-$k$ row.
We caught it because those two rows were the closest pair in Table~\ref{tab:t1} on all three benchmarks; a pairwise overlap check over selected frame \emph{indices} then showed 98.5\% agreement.
What is worth reporting is not the bug but how completely aggregate accuracy hid it.
Correcting the depth changes roughly 99.5\% of the selected frames on every benchmark, yet the aggregate moves by only $-1.05$ points on LongVideoBench and $+0.07$ on LVBench, while moving $+2.74$ on Video-MME.
On LVBench a reader comparing accuracies alone would see two-hundredths of a point between a faithful implementation and one selecting entirely different frames for 99.5\% of questions.
A single-number comparison cannot detect this class of error, and neither can a per-arm sanity check that only asks whether each row looks individually plausible.
What did detect it was comparing arms against \emph{each other}: if two rules that optimise different objectives select nearly the same frames, at least one is not implementing its objective.
We now run that overlap check by construction, and we recommend it wherever a table's rows are meant to be distinct methods.
The corrected row also reproduces the published Video-MME gain to within 0.07 points against a 2.81-point gap under the bug, which is evidence both that the correction is right and that this harness can match a published number when the port is faithful.

The disagreement is itself the measurement.
Two carefully controlled harnesses, running the same published rules at the same budget with the same encoder, differ by as little as 0.07 and as much as 3.74 points, a range that covers most of the gaps published selector comparisons are built on.
Two harnesses cannot bound the disagreement in general, but they are enough to show that cross-paper selector deltas of this size are fragile, including the numbers this paper's own argument targets, and it is why every claim we make is a paired contrast inside one harness rather than a comparison against a published figure.
It also means Table~\ref{tab:published} should be read as a measurement of harness disagreement, not as a reproduction verdict on any of these methods.

\paragraph{Joint allocation systems deserve decomposed reporting.}
This decomposition sharpens how to read systems such as Q-Frame, LDDR, and DAFS~\citep{zhang2025qframe,chen2026lddr,wang2026dafs}, whose end-to-end gains combine several causal routes at once.
Our controls find no value in a simple priority-proportional resolution rule, a positive effect from reconstructed LDDR-style importance, and the clearest downstream gain from temporal reinvestment: three different answers that a single end-to-end number would have blurred into one.
Future methods would be considerably easier to build on if they reported separate timestamp, resolution, and frame-count interventions alongside the joint result.

\paragraph{The mechanism suggests where to look next.}
OMP appears to work here in a way its original derivation does not predict.
The text query is nearly orthogonal to the frame-embedding cone, so the residual barely shrinks and sparse reconstruction never really happens; what remains useful is relevance plus decorrelation.
This has two consequences.
Late picks can extend multi-scene coverage after the text residual flattens, which is a genuine benefit.
But unconstrained diversity also rewards irrelevant novelty, which is the most frequent tag in our purposive failure audit.
Relevance floors, event-boundary constraints, and scorers with a smaller modality gap are the natural next interventions, and the residual trace gives a cheap way to check whether a proposed scorer has more query signal to offer before committing to a full evaluation.

\paragraph{Conclusions should span durations and answerers.}
Selection remains a property of the answerer--benchmark pair.
InternVL3-8B is nearly insensitive to OMP on medium and long Video-MME yet gains over ten points on LVBench, and OMP's advantage over uniform sampling is exactly zero at 15\,s.
A claim about long-video keyframe selection measured on one benchmark at one duration with one answerer is not safe to generalize, and our own results would have supported at least three different headlines depending on which cell we had chosen to report.

\section{Limitations}
\label{sec:limitations}

\paragraph{Two selectors are controlled reproductions, not full systems.}
FOCUS$^\star$ replays the published selection schedule against dense LongCLIP scores instead of reproducing the original budgeted ITM scorer, so its row tests the schedule under a matched scorer and not the FOCUS pipeline's accuracy or efficiency.
LDDR-select is stage 1 only, and the stage-2 allocation contrast is reconstructed from the paper rather than run from the authors' code.
Neither row should be read as a reproduction result for the original method.

\paragraph{Equivalence margins were chosen after seeing the data.}
The Qwen result covers pooled LongVideoBench 600 and 3600\,s examples within a $\pm3$-point margin and misses $\pm2$. Because the margin was selected after inspecting the interval rather than fixed in advance, it answers ``which margin does this sample clear'' rather than ``does the effect fall inside an independently justified bound''. We therefore report the interval and both margins and treat the result as descriptive; a confirmatory equivalence claim would need a pre-specified margin and a fresh replication. The tests are also uncorrected for multiplicity.
Video-MME compression has no paired test, and the small LVBench differences are descriptive rather than equivalence evidence.
The selector-by-budget interaction is underpowered for effects near one point, which is the scale at which it would matter.

\paragraph{Mechanism evidence is encoder-specific.}
The residual geometry and Gram statistics come from frozen LongCLIP stem embeddings, mostly on LongVideoBench-600\,s.
The selector ordering survives a swap to SigLIP (\S\ref{sec:scorer}), but both encoders share a contrastive image--text objective; BLIP-ITM heads, MLLM-attention scorers, and subtitle-aware selection remain untested, and the modality-gap explanation in \S\ref{sec:mechanism} is a property of this embedding space rather than of pursuit selection in general.
The model-assisted failure audit examined selected rather than randomly sampled failures, so its counts are exploratory and cannot estimate population rates.

\paragraph{Some secondary results were measured in a second compute environment.}
Tables~\ref{tab:t1}--\ref{tab:t3} and every claim drawn from them come from a single stack: RTX PRO 4500 GPUs with PyTorch 2.11/CUDA 13.0.
The selector-by-budget interaction in \S\ref{sec:reinvest}, the allocation-shape contrast in \S\ref{sec:compression}, and the variant sweep in Appendix~\ref{app:negatives} were measured on L40S GPUs with PyTorch 2.6.
In each case every arm of the contrast---including a re-run of the full-resolution baseline---was executed inside that second environment, so no reported difference is computed across the two stacks.
Replicating the baseline in both also lets us quantify the drift rather than assume it is small: the uniform eight-frame full-resolution arm reads 0.36 points higher on the second stack in the 3600\,s bin and is identical in the 600\,s bin, while the corresponding OMP arm reads 0.71 points lower.
Absolute accuracies from these sections are therefore not comparable to Tables~\ref{tab:t1}--\ref{tab:t3} and we never place them in the same table; the paired differences within each section are unaffected.

\paragraph{Evaluation scope is incomplete.}
Subtitles are disabled throughout, which leaves subtitle-anchored questions partly inaccessible to every visual selector and depresses all arms together.
GPT-5-mini compression uses a pixel ratio rather than Qwen token accounting, and some GPT selector results survive only as aggregates rather than per-item records.
Our rule set is also narrower than the comparison table it is matched against: MDP3 and Q-Frame appear in \citet{chen2026lddr} and we do not run them.
Neither has a published number at our exact cell---Qwen3-VL-8B, eight frames, LongCLIP scorer---so neither port could be validated against anything, which is the condition that let the AKS padding error stand undetected (\S\ref{sec:discussion}).
We judged two more unvalidated rows worse than a stated gap, and note that MDP3's relevance--diversity--sequentiality objective falls inside the DPP family whose behaviour \S\ref{sec:mechanism} already characterises, while Q-Frame couples selection to resolution and so measures the composite this paper decomposes rather than a rule at fixed budget.

\section{Conclusion}
\label{sec:conclusion}

In our matched harness, which frames reach the answerer is the largest of the three allocation decisions we varied.
Eight well-chosen frames beat sixteen uniformly spaced ones in LongVideoBench's hour-long bin, and OMP, run unmodified, is enough to get that result, outperforming uniform sampling by 5.7 to 11.8 points and staying within one point of a purpose-built modern selector.
Roughly half of the per-frame spatial budget can then be removed with little observed change, bounded within a post-hoc $\pm3$-point interval on pooled LongVideoBench long videos, and reinvesting those tokens in additional keyframes returns a further two to three points.
How far any of this generalises beyond this harness is exactly what the evaluation-sensitivity results above bound.
Select relevant frames first, compress them second, and use the savings to see more of the video, while remembering that how much any of this helps depends on the video's length, the benchmark, and the model doing the reading.

\section{Ethical Considerations}
\label{sec:ethics}

The experiments use existing public video-QA benchmarks and frozen open or API models.
No new human subjects, personal data, videos, or annotations were collected.
Question stems, but not answer options, enter the selector.
Benchmark videos remain governed by their original licenses and are not redistributed with the released artifacts.

\section{Data and Code Availability}

Evaluation code, selector implementations, analysis scripts, the selected frame indices for every rule and budget, and per-item predictions permitted by the benchmark licenses are available at \url{https://github.com/codeprakhar25/omp-keyframe-sampling}.
The release includes the scorer-swap pipeline of \S\ref{sec:scorer}, the fused-query arms of \S\ref{sec:method}, the script that builds Table~\ref{tab:published} from the cited papers' reported numbers, and the pre-registration documents for the registered arms, including the 3600\,s replication of \S\ref{sec:method}.
The release excludes benchmark video files and API credentials.

\bibliography{references}

\appendix

\section{FOCUS$^\star$ schedule replay}
\label{app:focus}

Every row in Table~\ref{tab:t1}, including FOCUS$^\star$, consumes the same dense 1\,fps LongCLIP stem-score vector.
The original FOCUS method instead treats scoring itself as a budgeted online process, in which clip-arm pulls purchase ITM evidence~\citep{zhu2025focus}.
Our replay preserves the temporal clip schedule and default hyperparameters but reveals already-computed LongCLIP scores to the bandit.
It therefore tests the schedule under a matched scorer, and says nothing about the efficiency or accuracy of the full FOCUS pipeline.

The replay uses 16\,s clips, three coarse pulls per clip, $\alpha{=}0.25$, a pull budget equal to half the candidate pool, and approximately $k/4$ surviving clips.
Unobserved rewards are interpolated from the nearest observed frame, and the random seed is fixed per item.
This boundary is why the row carries a star, and why we draw no conclusion from it about FOCUS under its native ITM protocol.

\section{Question categories do not define a stable selector regime}
\label{app:qtype}

LongVideoBench provides question-category tags, which invite the hypothesis that temporally referred questions need different frame selection from the rest.
We test it by comparing OMP with uniform sampling on the temporally referred categories (the \texttt{T*} group) against all remaining categories, without relabeling the data.

\begin{table}[h]
\centering
\small
\resizebox{\columnwidth}{!}{%
\begin{tabular}{llrccc}
\toprule
Bin & Slice & $n$ & Uniform & OMP & $\Delta$ (McNemar) \\
\midrule
600\,s & temporal \texttt{T*} & 148 & .5608 & .5811 & ${+}2.03$, $p{=}.74$ \\
600\,s & other & 264 & .5492 & .6591 & ${+}10.98$, $p{=}5.2\times10^{-4}$ \\
3600\,s & temporal \texttt{T*} & 222 & .4369 & .5270 & ${+}9.01$, $p{=}.013$ \\
3600\,s & other & 342 & .4942 & .5585 & ${+}6.43$, $p{=}.021$ \\
\bottomrule
\end{tabular}}
\caption{OMP minus uniform on official LongVideoBench categories. Percentage-point gains differ by duration, and the category interaction is not significant in either long bin.}
\label{tab:qtype}
\end{table}

The temporal slice looks resistant to selection at 600\,s and responsive at 3600\,s, which is the shape of a result that would be easy to over-interpret.
It is not one.
The category interaction is not significant in either bin, and which slice ``wins'' reverses between them.
Many \texttt{T*} stems refer to subtitles or speech while our selector is vision-only and subtitles are disabled, which offers one plausible source of the instability.
Two significant-versus-non-significant subgroup tests are not an interaction, so we infer no question-type router from these results.

\section{Selector variants}
\label{app:negatives}

\begin{figure}[t]
\centering
\includegraphics[width=\columnwidth]{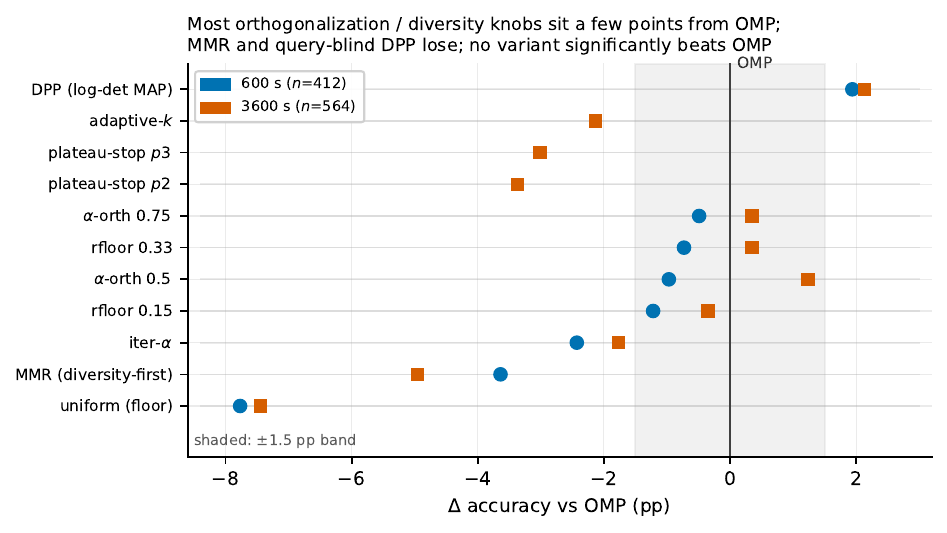}
\caption{Coverage-gated selector variants against OMP at $k{=}8$ on the 600 and 3600\,s LongVideoBench bins. The shaded $\pm1.5$-point band is a visual aid, not a statistical criterion.}
\label{fig:negatives}
\end{figure}

This sweep changes only the subset rule, on the same LongCLIP cache.
MMR underperforms; query-blind DPP ($\beta{=}0$) collapses toward uniform sampling, which is what a diversity term carrying no query signal would look like in this one configuration; and query-weighted DPP, residual floors, and partial orthogonalization all sit in a band around OMP.
No tested DPP variant significantly exceeds OMP in either long bin.

This sweep ran in the secondary compute environment described in \S\ref{sec:limitations}; its arms are internally paired against each other and are never compared against Tables~\ref{tab:t1}--\ref{tab:t3}.

These nulls do not show that all diversity objectives are equivalent.
They show something more specific and more useful: within this scorer geometry and budget range, moving around inside the relevance--diversity family produces less change than replacing uniform sampling or reallocating pixels into frames.
That is consistent with the residual analysis in \S\ref{sec:mechanism}: if the query-directed signal is exhausted after roughly five picks, then rules that differ only in how they trade relevance against diversity have little left to differ about.

\section{Exploratory failure audit}
\label{app:failure}

Two independent model-assisted visual passes inspected OMP failures at $k{=}8$ through a frame viewer: 41 cases spanning 16 categories in the 600\,s bin, and 52 cases spanning 17 categories in the 3600\,s bin.
Cases were chosen to cover categories and to include both OMP-only and shared failures; the sample is not random.
Reviewers compared the selected timestamps, the rendered frames, the question, the gold answer, and the model's prediction.
Primary tags were missed evidence, temporal offset, answerer error despite adequate frames, and questions not answerable from visual frames at all.

The most frequent tag in both passes was visually distinctive but irrelevant content consuming several of the eight picks: unrelated chapters in compilations, separate storylines in long narrative video, title and subscribe cards, and dark transitions. Because the sample is failure-conditioned, category-balanced by construction, model-assisted, and scored without blinding, inter-rater agreement, or a matched non-OMP control, we cannot call this pattern dominant, OMP-specific, or prevalent.
The opposite failure showed up on sequence-of-scenes and cross-scene tracking questions, where dense top-$k$ sampling covered one relevant moment but omitted another required scene.
The audit therefore motivates relevance-constrained diversity as a design direction.
It cannot estimate population failure rates, and it cannot establish that adding frames benefits any particular question category.

\end{document}